\documentclass{article}

\usepackage[preprint]{neurips_2023}

\usepackage[utf8]{inputenc} 
\usepackage[T1]{fontenc}    
\usepackage{hyperref}       
\usepackage{url}            
\usepackage{booktabs}       
\usepackage{amsfonts}       
\usepackage{nicefrac}       
\usepackage{microtype}      
\usepackage{xcolor}         
\usepackage{graphicx}
\usepackage{amsthm}
\usepackage{amssymb}
\usepackage{multirow}

\title{Disentangling Node Attributes from Graph Topology for Improved Generalizability in Link Prediction}

\author{%
  Ayan Chatterjee \\
  Network Science Institute\\
  Northeastern University\\
  Boston, MA 02115 \\
  \texttt{chatterjee.ay@northeastern.edu} \\
  \And
  Robin Walters \\
  Khoury College of Computer Sciences \\
  Northeastern University\\
  Boston, MA 02115 \\
  \texttt{r.walters@northeastern.edu} \\
  \AND
  Giulia Menichetti \\
  Brigham and Women’s Hospital \\
  Harvard Medical School \\
  Boston, MA 02115\\
  \texttt{giulia.menichetti@channing.harvard.edu} \\
  \And
  Tina Eliassi-Rad \\
  Khoury College of Computer Sciences \\
  Northeastern University\\
  Boston, MA 02115 \\
  \texttt{t.eliassirad@northeastern.edu} \\
}

\begin{document}

\maketitle

\begin{abstract}
    
    Link prediction is a crucial task in graph machine learning with diverse applications. We explore the interplay between node attributes and graph topology and demonstrate that incorporating pre-trained node attributes improves the generalization power of link prediction models. Our proposed method, \emph{UPNA} (Unsupervised Pre-training of Node Attributes), solves the inductive link prediction problem by learning a function that takes a pair of node attributes and predicts the probability of an edge, as opposed to Graph Neural Networks (GNN), which can be prone to topological shortcuts in graphs with power-law degree distribution. In this manner, \emph{UPNA} learns a significant part of the latent graph generation mechanism since the learned function can be used to add incoming nodes to a growing graph. By leveraging pre-trained node attributes, we overcome observational bias and make meaningful predictions about unobserved nodes, surpassing state-of-the-art performance (3$\times$ to 34$\times$ improvement on benchmark datasets). \emph{UPNA} can be applied to various pairwise learning tasks and integrated with existing link prediction models to enhance their generalizability and bolster graph generative models.

\end{abstract}

\section{Introduction}

Graph datasets are prevalent in various domains, such as friendship networks \citep{BALL_2013}, collaboration networks \citep{Wang2020}, protein interaction networks \citep{Qi2006}, power grids \citep{power_grid}, and transportation networks \citep{Lordan2020}. These real-world graphs are often sparse and partially observed, making link prediction for unobserved links an important problem \citep{LibenNowell2007}. Link prediction appears in many applications from predicting protein interactions \citep{Kovcs2019}, to exploring drug responses \citep{Stanfield2017}, to recommending products \citep{https://doi.org/10.48550/arxiv.2102.09185}, to completing knowledge graphs \citep{Nickel_2016}, and suggesting friends in social networks \citep{Adamic2003}.

Link prediction has been extensively studied, resulting in various methods including similarity-based indices, probabilistic approaches, and dimensionality reduction techniques \citep{Kumar2020}. For link prediction, latent representations such as Node2Vec \citep{node2vec} are commonly employed, capturing graph topology in low-dimensional feature vectors \citep{cao2015grarep,Perozzi_2014,Tang_2015,Wang2016}. Recent research highlights the significance of both graph topology and node attributes in link prediction \citep{ai_et_al_2022}.

Most existing link prediction approaches and benchmarks focus on transductive scenarios, where the train and test graphs share the same set of nodes \citep{wang2019dgl,benchamrking_jmlr_dgl,ogb}. However, transductive link prediction is primarily driven by the training graph's topology and biased by the observed graph's degree distribution \citep{Bonner_2022,ai-bind}. Similar observations have been made in node classification tasks, where label propagation often achieves comparable or superior performance compared to deep models \citep{benson_shortcuts}. Real-world applications often require semi-inductive and inductive link predictions on newly observed nodes, necessitating learning from node attributes independently of the graph topology. Existing models like GraIL \citep{grail} encode neighborhood information. These methods cannot deal with isolated nodes. To make accurate predictions in semi-inductive and inductive scenarios, a latent graph generation model without observational bias is crucial. These scenarios resemble the cold-start problem in recommendation systems, where suggesting a new product to an existing user resembles a semi-inductive setting, and predicting links with both unseen nodes resembles an inductive setting \citep{coldstart,ai-bind,Menichetti2022}. Inductive link prediction in knowledge graphs has gained significant attention recently \citep{ILPC}.

Temporal networks necessitate a broader understanding of dynamic complex systems, surpassing the limitations of static graphs \citep{holmes_2013}. Analyzing networks in a static manner can yield misleading insights into spreading processes and community structures \citep{Scholtes2016}. To comprehend the intricate dynamics of processes on networks, it is essential to uncover the evolution mechanism of temporal graphs. In discrete-time systems, newly arrived nodes connect to the previous temporal snapshot of the graph, creating semi-inductive or inductive link prediction scenarios \citep{holmes_2013, Scholtes2016}.

Identifying useful node attributes is crucial for enhancing node classification, link prediction, graph data augmentation, and graph generative models \citep{10.5555/3524938.3525387,simonovsky2018graphvae,shi2020graphaf}. Deep adversarial learning and variational auto-encoder-based approaches have been proposed to leverage real-world attributes for improving these tasks \citep{NAGG}. However, there is a lack of understanding regarding the interplay between graph topology and node attributes and the need for a formal quantification of attribute quality to effectively enhance and generalize these downstream tasks.

\textbf{Contributions:} \textbf{(1)} We address the issue of observational bias and poor performance on low-degree and unobserved nodes in transductive link prediction, highlighting the influence of topological shortcuts. \textbf{(2)} We emphasize the significance of learning the latent graph generation mechanism from node attributes to achieve accurate predictions for newly arrived isolated nodes. \textbf{(3)} We quantify the generalization power of link prediction models by assessing their inductive test performance and establishing the relationship with node attribute information. \textbf{(4)} We propose \emph{UPNA} (Unsupervised Pre-training of Node Attributes), a method that improves link prediction generalizability by training on large corpora independent of the training graph. We validate our approach on both static and time-evolving graphs.

\section{Problem Formulation}

\subsection{Static Graphs}

Consider a graph instance $G=(V,E,X)$, where $V$ represents the set of vertices (or nodes), $E$ represents the set of edges (or links), and the node attributes are captured in the matrix $X$. We focus on undirected unipartite graphs, although this formulation can be extended to encompass directed, bipartite, and multilayered graphs as well. For instance, $G$ could represent a protein-protein interaction network, where nodes correspond to proteins, links represent interactions between proteins, and node attributes are molecular structure embeddings obtained using ProtVec \citep{Asgari_2015}.

We construct a link prediction model $w$ using supervised learning on the set of links $E$. The edge set is partitioned into observed and unobserved edges during training as $E = E_{o} \cup E_{u}$. Our goal is to learn a function that maps the observed nodes and node attributes $(V_{o}, X_{o})$ to the observed edges $E_{o}$, with the hope that it will generalize to the unobserved edges $E_{u}$. We also define three types of link prediction scenarios based on the observed and unobserved nodes, denoted as $V_{o}$ and $V_{u}$, respectively:

\begin{itemize}
\item Transductive: Predicting $(a,b) \in E_{u}$, where $a, b \in V_{o}$,
\item	Semi-inductive: Predicting $(a,b) \in E_{u}$, where $a \in V_{o}$ and $b \in V_{u}$ or vice-versa,
\item	Inductive: Predicting $(a,b) \in E_{u}$, where $a, b \in V_{u}$.
\end{itemize}

In this work, we explore the semi-inductive and inductive link prediction scenarios. The link prediction model takes input $\{ V_o, X_o, E_o\}$, and makes predictions on $E_u$ induced by $V_u$.  

\begin{figure}[ht]
    \begin{center}    \includegraphics[clip,angle=0,width=\textwidth]{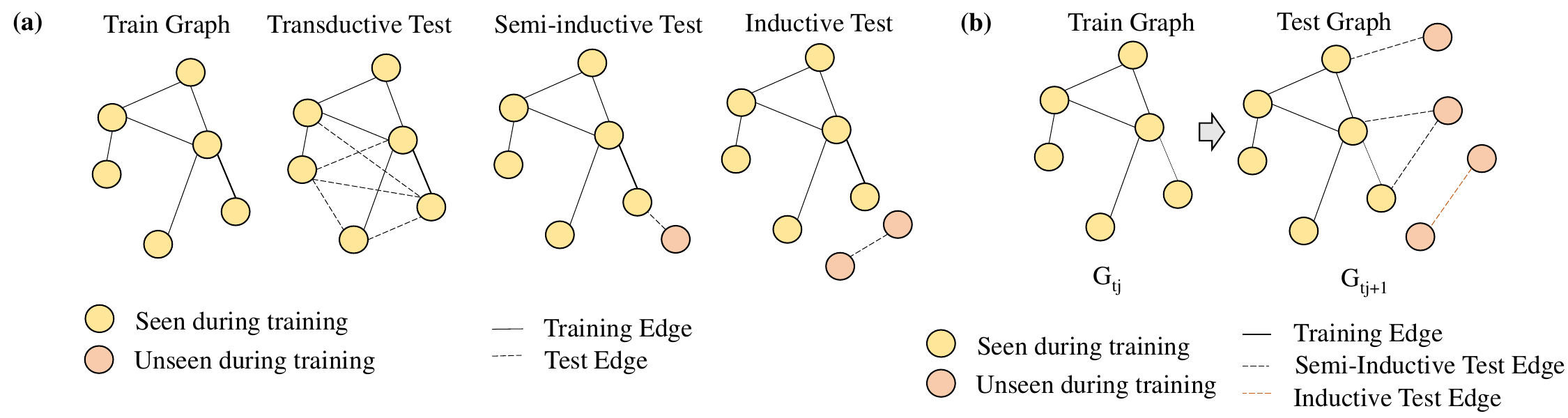}
    \end{center}
    \caption{\textbf{(a)} Transductive, semi-inductive, and inductive link prediction tasks in static graphs. \textbf{(b)} Semi-inductive and inductive link prediction tasks in discrete temporal graphs.}
\end{figure}

\subsection{Temporal Graphs}

\label{temporal_setting}

We consider discrete-time temporal graphs \citep{holmes_2013}, where the graph instances are captured at intervals of $\Delta t$, with time stamps $t_0$, $t_1 = t_0 + \Delta t$, $...$, $t_n = t_0 + (n-1) \Delta t$.

The temporal graph instances of the evolving network are denoted as $G_{t_0} = (V_{t_0}, E_{t_0}, X_{t_0})$, $G_{t_1} = (V_{t_1}, E_{t_1}, X_{t_1})$, $...$, $G_{t_n} = (V_{t_n}, E_{t_n}, X_{t_n})$, corresponding to time stamps $t_0$, $t_1$, $...$, $t_n$, respectively. We define the unobserved node set from time $t_j$ to $t_{j+1}$ as $V_u = V_{t_{j+1}} \setminus V_{t_{j}}$, and the unobserved edge set as $E_u = E_{t_{j+1}} \setminus E_{t_{j}}$. In this setting, the link prediction model takes input $G_{t_j} = (V_{t_j}, E_{t_j}, X_{t_j})$ and makes predictions on $E_u$ induced by $V_u$.

\section{Topological Shortcuts}

The influence of observation bias on link prediction in static graphs is significant. We demonstrate that state-of-the-art models heavily rely on graph topology, which can be replicated by simpler non-machine learning algorithms. Our experiments on Open Graph Benchmark (OGB) datasets \citep{ogb} reveal that top deep models' performance on the ogbl-ddi link prediction benchmark (drug-drug interaction network) can be matched by traditional configuration models (such as the traditional configuration model \citep{network_science_barabasi} and unipartite duplex configuration model \citep{Menichetti2014_weighted}). These simpler models ignore node attributes and rely solely on the degree sequence of the training graph, indicating a dependence on topological shortcuts while disregarding node attributes. However, such reliance on graph topology hinders generalization to unseen nodes in semi-inductive and inductive tests. Chatterjee et al. \citep{ai-bind} demonstrated that even a duplex bipartite configuration model, using only degree sequences of proteins and ligands in the training drug-target interaction network, achieves comparable test performance in predicting transductive protein-ligand interactions compared to state-of-the-art deep neural networks \citep{Huang2020,Huang2020_MolTrans}. Similar observations have been made for block-approximated exponential random graph models \citep{Adriaens2020,Mara2022} and simple link prediction heuristics \citep{Mara_2020,10.1145/3298689.3347058}.



Table \ref{table:config_model} compares the performance of traditional and duplex configuration models with three top-performing models: Adaptive Graph Diffusion Networks (AGDN, \citep{AGDN}), Path-aware Siamese Graph Neural Network (PSG, \citep{PSG}), and Pairwise Learning for Neural Link Prediction (PLNLP, \citep{PLNLP}) in the transductive setting using the OGB-provided benchmark train-validation-tests split. Notably, the configuration models outperform the state-of-the-art neural network models. A recent work \citep{stolman2022sdm} demonstrates that classic graph structural features outperform graph embedding-based methods in community labeling, specifically in terms of Hits@Top K. Similar observations regarding topological shortcuts are made in the Deep Graph Library (DGL) benchmark \citep{dgl} as well (\href{https://github.com/ChatterjeeAyan/UPNA}{here}). These shortcuts also manifest in other performance metrics such as AUROC and AUPRC \citep{ai-bind}.

\begin{table}[ht]
    \caption{The traditional and duplex configuration models outperform state-of-the-art neural network models in Hits@Top K on the ogbl-ddi dataset, using K=20 as recommended by the OGB benchmark.}
    \label{table:config_model}
    \begin{center}
    \begin{tabular}{l l l l}
    \hline
    \multicolumn{1}{c}{\bf Model} &
    \multicolumn{1}{c}{\bf Hits@Top K(\%)}\\
    \hline
    \hline
    Traditional Configuration Model & $\mathbf{0.99 \pm 0.00}$ \\
    \hline
    Duplex Configuration Model & $\mathbf{0.99 \pm 0.00}$ \\
    \hline
    AGDN  & $0.95 \pm 0.01$ \\
    \hline
    PSG & $0.93 \pm 0.01$ \\
    \hline
    PLNLP & $0.91 \pm 0.03$ \\
    \hline
    \end{tabular}
    \end{center}
\end{table}

\subsection{Degree Bias in Topological Shortcuts}

\begin{figure}[ht]
    \begin{center}    \includegraphics[clip,angle=0,width=\textwidth]{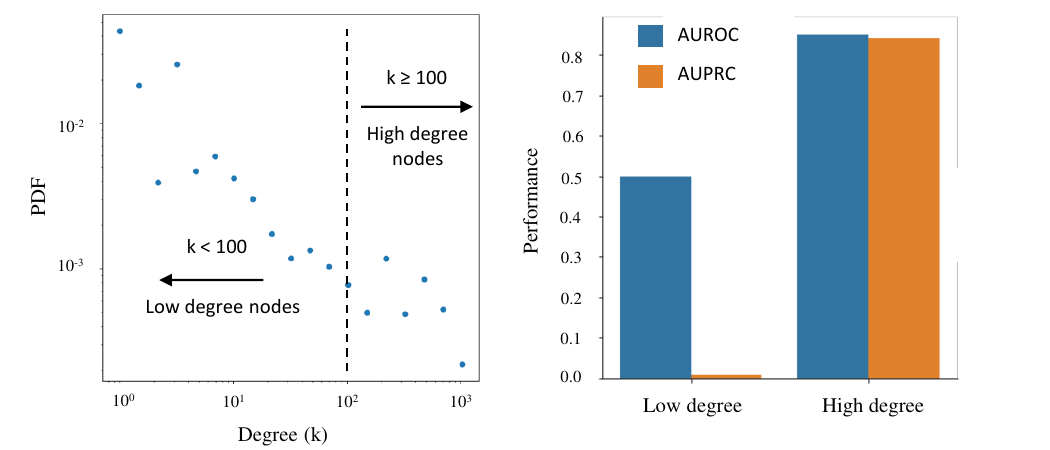}
    \end{center}
    \caption{The transductive test performance on the ogbl-ddi dataset varies across degree bins. Low-degree nodes, which have less influence on topological shortcuts, exhibit poorer test performance compared to high-degree nodes (hubs).}
    \label{fig:degree_bias}
\end{figure}

The presence of high-degree nodes (hubs) \citep{Barabasi1999} in power-law degree distributions creates a topological shortcut that influences link prediction performance. Link prediction models primarily learn from the degree information of hubs, resulting in accurate predictions for hub-related links and excellent overall test performance. However, this leads to poor performance for low-degree nodes, as shown in Figure \ref{fig:degree_bias} for the ogbl-ddi dataset using a multi-layer perceptron (MLP) model (Check \href{https://github.com/ChatterjeeAyan/UPNA}{here} for model description and reproducibility). In semi-inductive and inductive settings, where there is limited topological information, relying solely on topological shortcuts fails to make predictions for newly arrived nodes. Therefore, inductive tests unveil the true power of machine learning in link prediction, and understanding the latent graph generation mechanism becomes crucial in these scenarios, motivating the focus of this work.

\section{State-of-the-art Models in Inductive Link Prediction}

We investigate semi-inductive and inductive link prediction settings after addressing the limitations of transductive tests. We propose methods for inductive tests on the OGB link prediction benchmark datasets by using a random node split \citep{grail}. In this train-test-validation split method, we randomly split the nodes of the original graph $V$ into three groups ($V_{train}$, $V_{validation}$, and $V_{test}$) at an 80:10:10 ratio. Then, we obtain the subgraphs $G_{train}$, $G_{validation}$, and $G_{test}$ corresponding to $V_{train}$, $V_{validation}$, and $V_{test}$, respectively.

Our findings reveal that state-of-the-art link prediction models perform poorly in inductive tests, achieving lower performances than a simple MLP trained on the node attributes. The non-overlapping topological information between the train and test graphs in the inductive setting necessitates models to rely solely on node attributes to make accurate link predictions for unseen nodes during training.


\subsection{State-of-the-art Link Prediction Models Fail in Inductive Tests}

We evaluate the importance of node attributes in inductive link prediction using OGB-defined MLPs (\href{https://github.com/ChatterjeeAyan/UPNA}{here}). These MLPs take concatenated attributes of the two nodes at the end of each edge as input. We compare their performance with the state-of-the-art model PLNLP \citep{PLNLP} on the OGB link prediction leaderboard. In the case of ogbl-ppa, we use 58-dimensional one-hot feature vectors representing the associated species with proteins. For ogbl-collab, we use 128-dimensional features obtained by averaging word embeddings of authors' published papers. For ogbl-ddi, we utilize 300-dimensional Mol2vec \citep{Mol2vec} embeddings of drug structures. Our findings, summarized in Table \ref{tab:comparison_inductive}, demonstrate that the inductive performance of PLNLP is significantly lower compared to its transductive performance, while the MLP performs better in the inductive test scenario.


\begin{table*}[ht]
    \centering
    \caption{PLNLP performs worse in inductive tests compared to its transductive performance and is outperformed by an MLP trained solely on node attributes. Hits@Top K is evaluated with default values: K=100 for ogbl-ppa, K=50 for ogbl-collab, and K=20 for ogbl-ddi. We conduct a 5-fold cross-validation for each dataset. \\} 
	\label{tab:comparison_inductive}
	\begin{tabular}{l l l l}
        \hline
        {\bf Dataset} & {\bf PLNLP in Transductive Test} & {\bf PLNLP in Inductive Test} & {\bf MLP on node attributes} \\
		& {\bf Hits@Top K(\%)} & {\bf Hits@Top K(\%)} & {\bf Hits@Top K(\%)} \\
		\hline
        \hline
		ogbl-ppa & $32.38 \pm 2.58$ & $0.09 \pm 0.03$ & $0.39 \pm 0.03$ \\
        \hline
		ogbl-collab & $70.59 \pm 0.29$ & $11.56 \pm 0.93$ & $36.44 \pm 3.11$ \\
        \hline
		ogbl-ddi & $90.88 \pm 3.13$ & $0.01 \pm 0.02$ & $0.39 \pm 0.02$\\
		\hline
	\end{tabular}
\end{table*}


\section{Theoretical Background}

\label{theory_main}

\label{theory}

In this section, we delve into the theoretical foundations of our proposed node attribute engineering approach, aiming to enhance the performance of semi-inductive and inductive link prediction. Building upon the established framework of statistical learning theory \citep{books/daglib/0033642,xu_2017_neurips}, we formalize the performance of semi-inductive and inductive link prediction as the measure of generalization power exhibited by a link prediction model. To this end, we define the instance space, denoted as $E$, which represents the set of edges, and the hypothesis space, denoted as $W$, which encompasses the link prediction hypotheses. The loss function, denoted as $l: W \times E \longrightarrow \mathbf{R}^+$, is a non-negative function that quantifies the discrepancy between the predicted and true edges. Given an input of $n$ edges, represented as the $n$-tuple $E=(e_1, ..., e_n)$, which are assumed to follow an unknown distribution with mean $\mu$ characterizing the underlying graph generation process, we can express the population risk (i.e., the expected loss) of a link prediction hypothesis $w \in W$ as follows: 

\begin{equation}
    L_{\mu}(w) \triangleq \mathbf{E}[l(w,E)] = \int_{E} l(w,e) \mu(de)
\end{equation}

The parameter $\mu$ in the aforementioned formulation is closely associated with the underlying graph generation process. In the context of inductive link prediction, the generalization error quantifies the capability of the link prediction algorithm to capture and learn the graph generation mechanism based on the finite set of observed edges $E$. Specifically, the generalization error on $\mu$ can be assessed by examining the discrepancy between $L_{\mu}(w)$ and $L_E(w)$. Considering the expected value of this generalization error, we obtain the following expression:

\begin{equation}
    gen(\mu, P_{w|E}) \triangleq \mathbf{E}[L_{\mu}(w) - L_E(w)],
\end{equation}

which measures the gap between learning the latent generation mechanism and learning from the finite set of observed edges.

\begin{equation}
    \mathbf{E}[L_\mu(w)] = \mathbf{E}[L_E(w)) + gen(\mu, P_{w|E}]
\end{equation}

\textbf{Theorem 1:} If $L(w)$ is $\sigma$-subgaussian under $P_{E,w} = \mu^{\bigotimes n} \bigotimes P_{w|E}$, where $\bigotimes$ represents the product of two marginal distributions, then 

\begin{center}
    $|\mathbf{E}[L_\mu(w)] - \mathbf{E}[L_E(w)]| \leq \sqrt{2 \sigma^2 I(E;w)}$
\end{center}

where $I(E;w)$ quantifies the information shared between the edges and the information learned by the link prediction model. 

\emph{Proof: } Please refer to Appendix \ref{proof_th_1}.

Rewriting the above inequality in terms of the generalization error of the link prediction models, we get:

\begin{equation}
    |gen(\mu, P_{w|E})| \leq \sqrt{2\sigma^2I(E;w)}
    \label{eq:upper_bound}
\end{equation}

The mutual information $I(E;w)$ quantifies the shared information between the link prediction hypothesis and the training data. Specifically, the link prediction hypothesis $w$ consists of two distinct components: $w_{t}$, which utilizes the graph topology, and $w_{a}$, associated with the node attributes. Consequently, we can partition the mutual information in the following manner:

\begin{equation}
    I(E;w) = I(E;w_t,w_a) = I(w_t,w_a) - I(w_t,w_a|E)
\end{equation}

To minimize the generalization error of the link prediction model, we aim to minimize the mutual information $I(w_t,w_a)$ between node attributes and graph topology, independent of the training dataset $E$. This requires leveraging node attributes to improve semi-inductive and inductive link prediction performance. Unsupervised pre-training using unlabeled data has been shown to enhance generalization when labeled data is limited \citep{Blum1998}. Similarly, pre-training node attributes on a large corpus improves the generalizability of link prediction models on unseen nodes. This pre-training enables models to converge faster and achieve better generalization \citep{JMLR:v11:erhan10a}.



We measure information in different node attributes using unsupervised clustering and Davies-Bouldin score \citep{Davies1979} to identify attributes suitable for inductive link prediction. Adjusted mutual information is used to quantify shared information between graph topology and node attributes, validating the theoretical background developed in Section \ref{theory_main}.


\section{Experiments}

\subsection{Methodology}

Real-world graphs often exhibit community structures, representing groups with shared characteristics or functions. Social networks have communities based on interest groups, while protein-protein interaction networks organize proteins into communities based on their metabolic functions \citep{Radicchi2004}. Traditional community detection algorithms can identify communities formed by locally dense subgraphs \citep{Girvan2002,Rosvall2009}. Additionally, unsupervised clustering on node attributes aids in detecting communities beyond the graph's topology \citep{Yang_2013}.

\begin{figure}[ht]
    \begin{center}    \includegraphics[clip,angle=0,width=\textwidth]{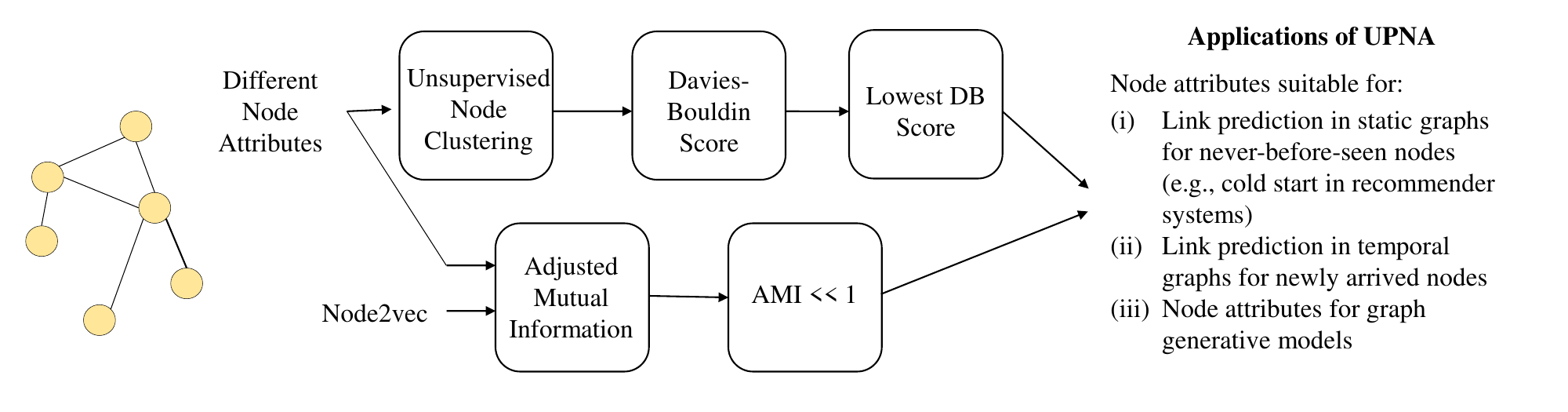}
    \end{center}
    \caption{UPNA methodology selects node attributes essential for various downstream applications by capturing the graph generation mechanism.}
    \label{fig:upna-flow}
\end{figure}

\emph{UPNA} utilizes k-means clustering \citep{Macqueen67somemethods} on node attributes to create unsupervised clusters. The quality of the clusters is evaluated using the Davies-Bouldin score \citep{Davies1979}, where lower scores indicate higher information content in the node attributes. Adjusted mutual information (AMI) \citep{Vinh2009} is measured between pre-trained attribute-based clusters and Node2Vec-based clusters to quantify $I(w_t,w_a)$ (Section \ref{theory}). Low AMI values indicate that the node attributes contain distinct information from the graph topology, making them suitable for inductive prediction. The \emph{UPNA} methodology is illustrated in Figure \ref{fig:upna-flow}.



We evaluate the quality of unsupervised node clusters using different node features: (a) Node2Vec encodes graph topology, (b) Pre-trained node attributes independent of the graph topology, (c) Randomly shuffled pre-trained node attributes for each node, and (d) Pre-trained node attributes replaced with random entries from a uniform distribution. Random and shuffled attribute versions provide insights into the relationship between attribute information and inductive link prediction performance.

\subsection{Observations on Inductive Link Prediction}

We apply the UPNA methodology to multiple link prediction benchmark datasets and compare it with state-of-the-art models in inductive link prediction. 

\subsubsection{Static Graphs}

We train the node attributes of the OGB graphs in an unsupervised manner. For ogbl-ppa, we use 100-dimensional ProtVec vectors \citep{Asgari_2015} trained on 546,790 amino acid sequences from the Swiss-Prot database \citep{Bairoch1996}. For ogbl-collab, we use 128-dimensional Word2Vec embeddings trained on approximately 6 billion tokens from Google news articles \citep{word2vec}. For ogbl-ddi, we use the pre-trained 300-dimensional Mol2vec embeddings \citep{Mol2vec} trained on a corpus of 19.9 million chemicals from ZINC \citep{Irwin2012} and ChEMBL \citep{Gaulton2011} libraries.

\begin{figure*}[ht]
    \begin{center}
    \includegraphics[clip,angle=0,width=\textwidth]{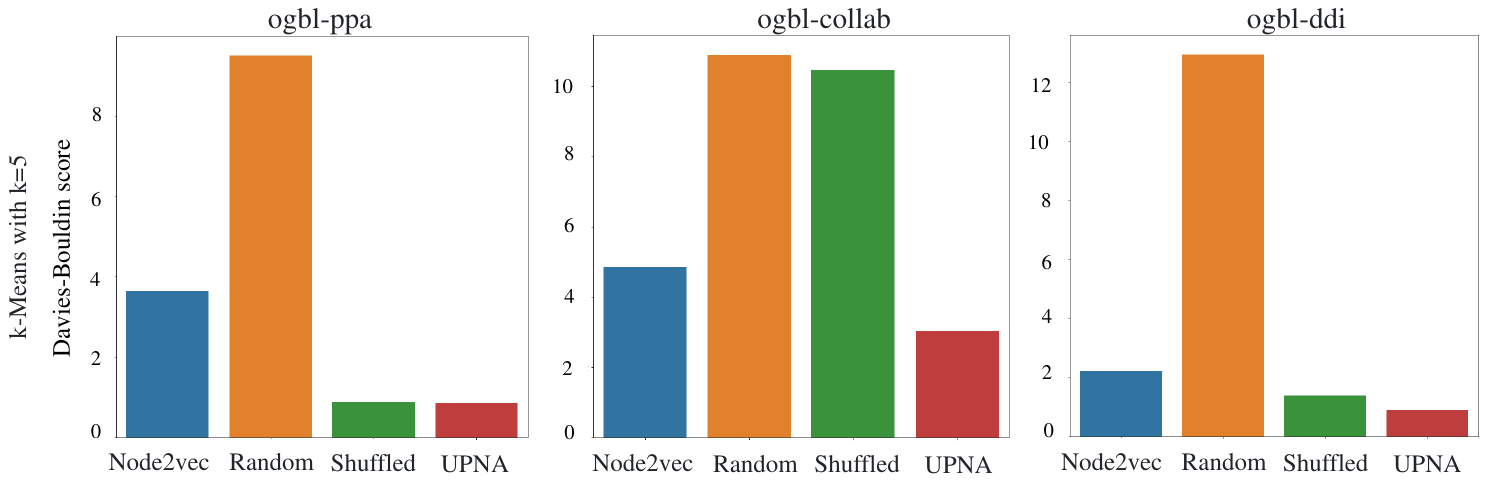}
    \end{center}
    \caption{Comparing the unsupervised clusters using Davies-Bouldin score on OGB link prediction benchmarks, the pre-trained node attributes consistently yield the lowest score. This indicates their suitability for inductive link prediction tasks across different graph datasets. The K-means algorithm with K=5 is employed, and similar observations hold for different K values.}
    \label{fig:DB_Score}
\end{figure*}

Comparing the Davies-Bouldin scores for different node attributes reveals that pre-trained node attributes achieve the lowest score, indicating more meaningful node clustering (see Figure \ref{fig:DB_Score}). This suggests that pre-trained node attributes contain valuable information beyond the graph topology, making them suitable for inductive tests. Additionally, Table \ref{table:AMI} shows low adjusted mutual information (AMI) between the pre-trained node attributes and Node2Vec, indicating limited shared information with the graph topology and low $I(w_t,w_a)$. This confirms that pre-trained node attributes capture minimal topological information from the training graph, enhancing the generalization power of link prediction models (Eq. \ref{eq:upper_bound}).

\begin{table}[ht]
    \caption{We compare the adjusted mutual information between the unsupervised clusters obtained using only Node2vec and the pre-trained node attributes. Node2vec uses the graph topology to create the node clusters. Low AMI values ($\ll 1$) indicate that the pre-trained attributes share minimal information with the graph topology.}
    \label{table:AMI}
    \begin{center}
    \begin{tabular}{l l}
    \hline
    \multicolumn{1}{c}{\bf Dataset}  & \multicolumn{1}{c}{\bf $I(w_t,w_a)$} \\
    \hline
    \hline
    ogbl-ppa & 0.17 \\
    \hline
    ogbl-collab & 0.08 \\
    \hline
    ogbl-ddi & 0.22 \\
    \hline 
    \end{tabular}
    \end{center}
\end{table}

\begin{figure*}[ht]
    \begin{center}
    \includegraphics[clip,angle=0,width=\textwidth]{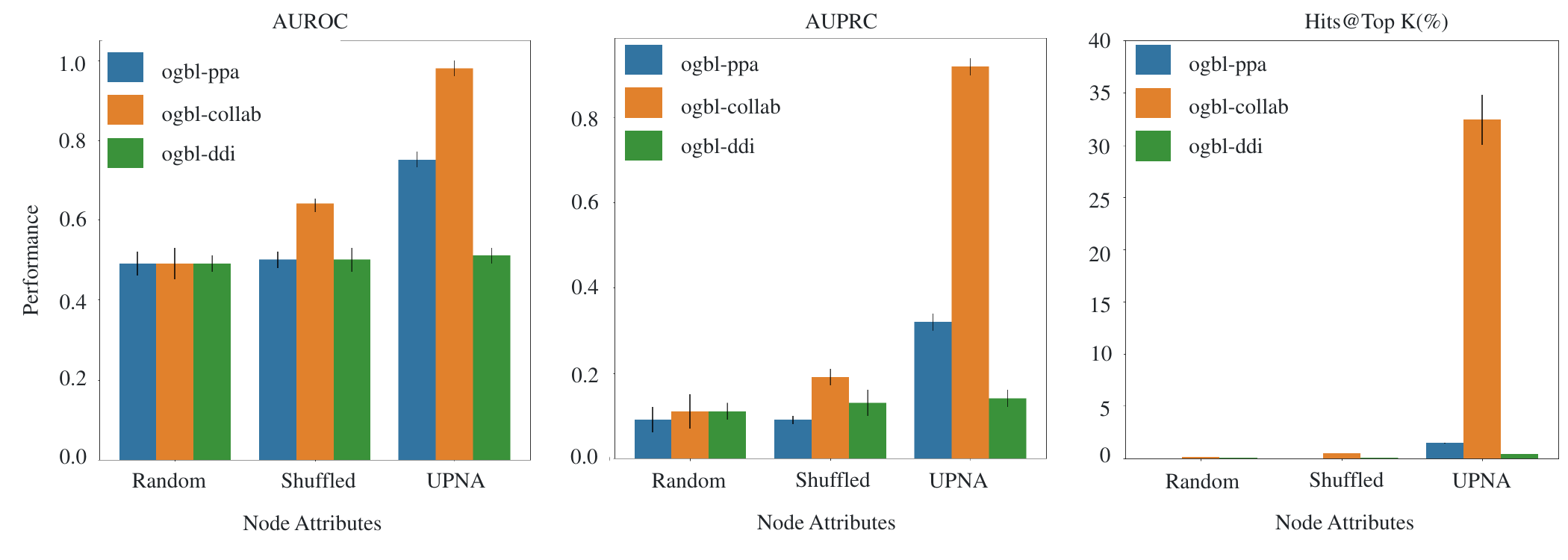}
    \end{center}
    \caption{We compare AUROC, AUPRC and Hits@Top K performances for different node attributes in inductive tests. The pre-trained node attributes show the highest performance in all of the OGB link prediction datasets.}
    \label{fig:Perf-pre-trained}
\end{figure*}

Using the aforementioned node features, we perform inductive link prediction on the OGB benchmark using the provided MLP architectures (\href{https://github.com/ChatterjeeAyan/UPNA}{here}). Figure \ref{fig:Perf-pre-trained} demonstrates that the pre-trained node attributes, which exhibit the lowest Davies-Bouldin scores and yield the best unsupervised node clustering, achieve superior performance in inductive link prediction.


\subsection{Temporal Graphs}

We evaluate UNAP on temporal networks for link prediction on newly arrived nodes (see Subsection \ref{temporal_setting}.2). We utilize the reddit hyperlink network dataset \citep{kumar2018community}, which spans 3 years (2014-2017) and consists of subreddit communities. The pre-trained node attributes are generated from post content using GloVe word embeddings trained on 840B words \citep{pennington2014glove}. For prediction, we employ a 3-layer MLP decoder with hidden layer sizes of 100, a learning rate of 0.001, 200 epochs, ReLU activation, and ADAM solver. The pre-trained node attributes with the lowest Davies-Bouldin score outperform others in predicting links for newly arrived nodes across different years. We also compare UNAP with DyHATR \citep{DyHATR}, a state-of-the-art temporal link prediction model, in inductive link prediction. Table \ref{tab:temporal_results} summarizes these findings.

\begin{table*}[ht]
  \centering
  \caption{We evaluate link prediction performance on newly arrived nodes across various temporal instances of the subreddit network. The performance is correlated with the Davies-Bouldin scores (DB) of node attributes (with $K$-means clustering, $K=5$). Lower scores indicate more informative attributes and higher prediction performance. Pre-trained node attributes yield the best performance. \\}
  
  \begin{tabular}{c l l l l l l l l l l l }
    \hline
    \multirow{2}{*}{Attributes} &
      \multicolumn{3}{c}{2014-2015} &
      \multicolumn{3}{c}{2016-2017} &
      \multicolumn{3}{c}{2017-2018} \\ 
      & {DB} & {AUROC} & {AUPRC} & {DB} & {AUROC} & {AUPRC} & {DB} & {AUROC} & {AUPRC} \\
        \hline \hline
        UPNA & \textbf{2.9} & \textbf{0.70} & \textbf{0.66} & \textbf{0.2} & \textbf{0.63} & \textbf{0.60} & \textbf{2.8} & \textbf{0.69} & \textbf{0.65} \\
        \hline
        Shuffled & 20.5 & 0.66 & 0.62 & 15.04 & 0.59 & 0.55 & 16.9 & 0.60 & 0.56 \\
        \hline
        Random & 21.9 & 0.5 & 0.5 & 22.3 & 0.5 & 0.5 & 21.3 & 0.51 & 0.51 \\
        \hline
        DyHATR & - & 0.45 & 0.25 & - & 0.45 & 0.48 & - & 0.46 & 0.28 \\
        \hline
  \end{tabular}
  \label{tab:temporal_results}
\end{table*}


\section{Related work}

Machine Learning models in link prediction utilize both graph topology and node attributes \citep{ai_et_al_2022}. In transductive tests, GraphSAGE \citep{graphsage} and similar methods perform well when train and test graphs have similar topologies and neighborhood information is known for newly arrived nodes. Random edge splits create similar topologies between train and test graphs, making basic topological features (e.g., degree, centrality) adequate for link prediction \citep{ghasemian:pnas2020}. 

Joachims \citep{joachims99} first highlighted the difficulty of inductive tests in various machine learning domains, including graph-based link prediction. Planetoid \citep{planetid}, GraIL \citep{grail}, and GraphSAGE \citep{graphsage} are notable methods for inductive link prediction that rely on neighborhood link information and exploit topological shortcuts. DEAL \citep{Hao_2020} combines topological information and node attributes for link prediction in both transductive and inductive settings. SEG \citep{ai_et_al_2022} employs a one-layer GCN for topology encoding and an MLP on node attributes, improving transductive performance. Recent challenges like ILPC \citep{ILPC} have spurred the development of state-of-the-art inductive models, such as CascadER \citep{CascadER} and Inductive NodePiece \citep{NodePiece}.

Node attributes play a crucial role in graph stream problems, dynamic network research \citep{jiang2015link,10.1145/3159652.3159674}, and cold-start scenarios \citep{5693971,10.1145/3446427}. While pre-training has been used in dynamic network research \citep{10.1145/3534678.3539396} and cold-start problems \citep{liu2023graphprompt,10.1145/3544107,10.1145/3555372}, it has primarily focused on training GNN model parameters rather than node attributes. Our method, \emph{UPNA}, introduces pre-training for node attributes using large corpora, enabling generalizable embeddings. By integrating \emph{UPNA} into existing models for link prediction in dynamic networks and cold-start problems, we enhance their generalizability.


\section{Conclusion and Future Work}

Our work derives motivation from the tendency of link prediction models to harness the train graph's topology, yielding outstanding performance in transductive tests. We have underscored the significance of inductive link prediction, devising a prediction framework for both static and temporal graphs. We have observed a notable decline in the performance of state-of-the-art link prediction models when confronted with the challenges of inductive tests. Astonishingly, their performance often pales in comparison to that of a simple MLP operating solely on the node attributes.

To provide a theoretical underpinning for inductive link prediction, we have delved into the intrinsic relationship between the shared information of node attributes and the graph topology. Additionally, we have crafted a method to quantify the quality of node attributes and conducted empirical experiments that demonstrate the exceptional suitability of pre-trained node attributes in enhancing generalizability for link prediction and capturing the intricate nuances of graph generation.

Moving forward, \emph{UPNA} can be extended by integrating pre-trained node attributes into multiple state-of-the-art link prediction models, aiming to achieve improved overall performance in link prediction. Moreover, we intend to incorporate the Davies-Bouldin score of node attributes into the pre-training objective function, facilitating the attainment of optimal pre-training outcomes. Lastly, delving into the impact of the training corpus size for pre-trained node attributes on inductive link prediction performance will further enrich our understanding of the hypotheses put forth by Erhan et al. regarding unsupervised pre-training \citep{JMLR:v11:erhan10a}.

In summary, \emph{UPNA} stands as a powerful tool capable of selecting the most fitting node attributes to enhance the generalizability of link prediction models for newly arrived nodes, while concurrently elevating the capabilities of graph generative models.\\

Our software and information about the data used in the experiments are at \url{https://github.com/ChatterjeeAyan/UPNA}.








\bibliographystyle{ACM-Reference-Format}
\bibliography{main}


%
\appendix


\section*{Appendix}

\section{Topological Sense Features}

\begin{table}[ht]
    \caption{In the context of ogbl-ddi, we examine the factors that significantly impact the test performance of link prediction. Our findings reveal that the node degree (represented by k) and the triangle count ($\Delta$) for each node emerge as the most influential contributors. Additionally, we consider the local clustering coefficient (CC) and the betweenness centrality ($C_B$) as higher-order network properties. To conduct our analysis, we utilize the established benchmark train-validation-test split from OGB, ensuring a standardized evaluation setup.}
    \label{tab:mlp_node_feat_perf}
    \begin{center}
    \begin{tabular}{l l l l}
    \hline
    \multicolumn{1}{c}{\bf Features} & \multicolumn{1}{c}{\bf AUROC} & \multicolumn{1}{c}{\bf AUPRC} \\
    \hline
    \hline
    k+CC+$\Delta$+$C_{B}$ & $1.0$ & $0.99$ \\
    \hline
    CC+$\Delta$+$C_{B}$ & $0.98$ & $0.98$ \\
    \hline
    CC+$C_{B}$ & $0.57$ & $0.62$ \\
    \hline
    \end{tabular}
    \end{center}
    \label{table:mlp_node_feat_perf}
\end{table}


\section{Proof of Theorem 1}

\label{proof_th_1}

\begin{proof} Using Corollary 4.15 in \citep{Boucheron2013} for the duality of the entropy of general random variables, we can write the following measure-theoretic bounds: 

\begin{equation}
    I(E;w) = \sup \{\int_{\mu} L(w) d\mu  - \log \int_E \exp({L(w)}) de \}
\end{equation}

Now, replacing the integrals with expectations and replacing the supremum with $\geq$, we get:

\begin{equation}
    I(E;w) \geq \mathbf{E} [\lambda L_\mu(w)] - \log \mathbf{E} [e^{\lambda L_E(w)}] 
    \geq \lambda (\mathbf{E} [L_\mu(w))] - \mathbf{E} [L_E(w)]) - \frac{\lambda^2 \sigma^2}{2},
    \label{eq:ineq_th}
\end{equation}

where $\lambda \in \mathbf{R}$ and the second step is derived from the subgaussian nature of $L(w)$ (see Appendix \ref{subgaussian_loss_function} for the empirical validation of the subgaussain nature of the loss function in link prediction):

\begin{equation}
    \log \mathbf{E} [e^{\lambda (L_E (w) - \mathbf{E} [L_E (w)]}] \leq \frac{\lambda^2 \sigma^2}{2}, \forall \lambda \in \mathbf{R}
    \label{eq:parabola}
\end{equation}

The inequality in Equation \ref{eq:ineq_th} generates a non-negative parabola in $\lambda$, which has a non-positive discriminant. Thus, we get:

\begin{equation}
    |\mathbf{E} [L_\mu (w)] - \mathbf{E} [L_E (w)]| \leq \sqrt{2 \sigma^2 I(E;w)}
\end{equation}

\end{proof}


\section{Random Node Split on OGB for Inductive Tests}

Our investigation reveals a notable discrepancy between the random node split and random edge split methods in terms of the number of lost edges. Specifically, the random node split approach exhibits considerably fewer lost edges. Leveraging this insight, we adopt the random node split method to construct inductive tests for the OGB link prediction datasets. To provide a comprehensive overview, we present the statistics regarding the number of nodes in Table \ref{table:random_node_split_1} and the number of edges in Table \ref{table:random_node_split_2} for each of the undirected link prediction datasets. These tables offer valuable insights into the dataset characteristics and facilitate a thorough understanding of the experimental setup.

\begin{table}[ht]
    \caption{Number of nodes in different OGB datasets for inductive tests.}
    \label{table:random_node_split_1}
    \begin{center}
    \begin{tabular}{l l l l}
    \hline
    \multicolumn{1}{c}{\bf Dataset} & \multicolumn{1}{c}{\bf Train Nodes} & \multicolumn{1}{c}{\bf Validation Nodes} & \multicolumn{1}{c}{\bf Test Nodes} \\
    \hline
    \hline
    ogbl-ppa & 461,031 & 57,629  & 57,629 \\
    \hline
    ogbl-collab & 188,694 & 23,587 & 23,587 \\
    \hline
    ogbl-ddi & 3,413 & 427 & 427 \\
    \hline 
    \end{tabular}
    \end{center}
\end{table}

\begin{table*}[ht]
    \caption{Edges in OGB link prediction datasets for inductive tests.}
    \label{table:random_node_split_2}
    \begin{center}
    \begin{tabular}{l l l l l}
    \hline
    \multicolumn{1}{c}{\bf Dataset} & \multicolumn{1}{c}{\bf Train Edges} & \multicolumn{1}{c}{\bf Validation Edges} & \multicolumn{1}{c}{\bf Test Edges} & \multicolumn{1}{c}{\bf Edges lost} \\
    \hline
    \hline
    ogbl-ppa & 19,460,915 & 296,195 & 303,563 & 10,265,600 \\
    \hline
    ogbl-collab & 821,974 & 12,739 & 12,826 & 437,926 \\
    \hline
    ogbl-ddi & 864,478 & 13,869 & 11,433 & 445,109 \\
    \hline
    \end{tabular}
    \end{center}
\end{table*}


\section{Subgaussian nature of the loss function in link prediction}

\label{subgaussian_loss_function}

\begin{figure}[b]
    \begin{center}    \includegraphics[clip,angle=0,width=0.7\textwidth]{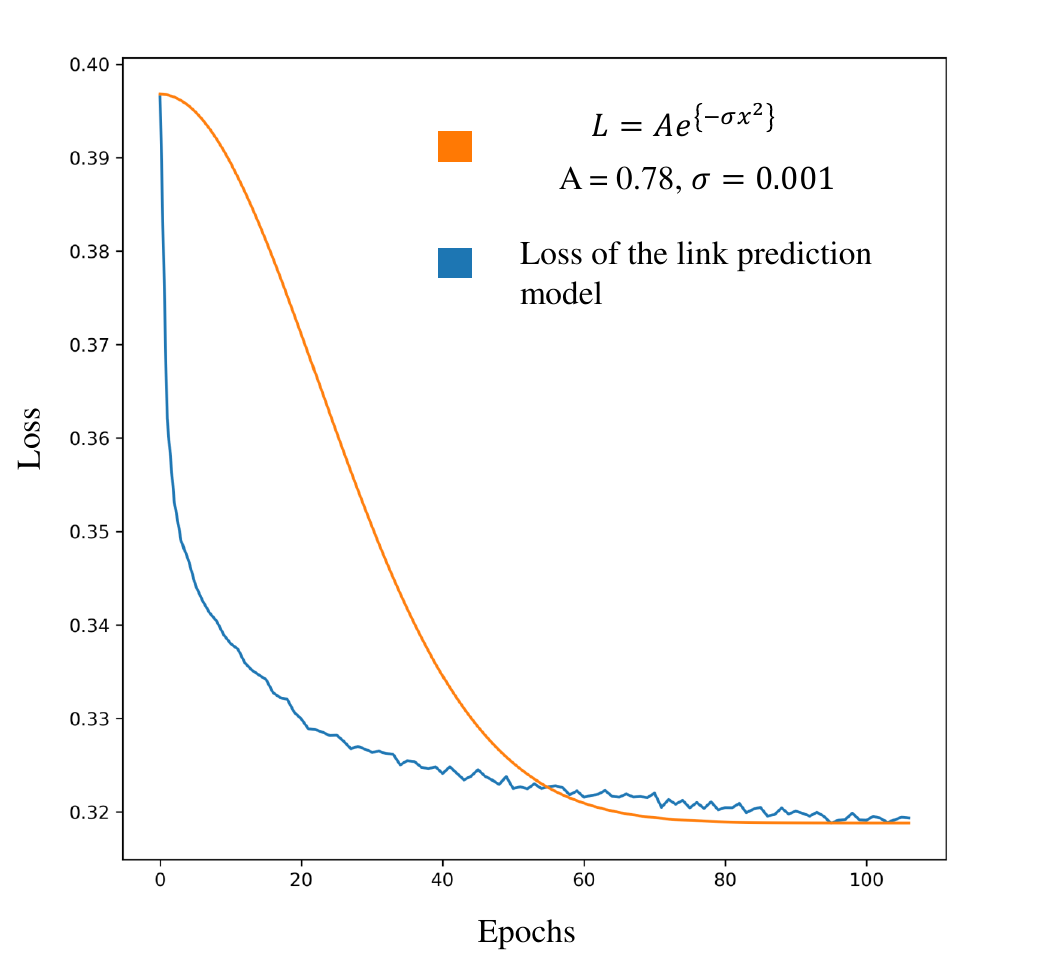}
    \end{center}
    \caption{The loss function in link prediction decays faster than a Gaussian tail, creating a subgaussian behavior.}
    \label{fig:subgaussian_loss}
\end{figure}

We delve into an empirical investigation of the subgaussian nature exhibited by the loss function employed in the link prediction model on the subreddit network. The subgaussianity of the loss function leads to the formulation of Eq. \ref{eq:ineq_th}, which, in turn, enables the derivation of the parabolic form depicted in Eq. \ref{eq:parabola}. Our analysis, as showcased in Figure \ref{fig:subgaussian_loss}, demonstrates that the loss function during the training phase of the link prediction model decays at a more rapid pace than the Gaussian tail characterized by $L = A e^{-\sigma x^2}$, where we determine $A=0.78$ and $\sigma=0.001$ through an exponential fit. Consequently, we can ascertain that the loss function exhibits a subgaussian nature with a parameter of $\sigma=0.001$.

\end{document}